\documentclass{article}
\usepackage{arxiv}  

\usepackage[utf8]{inputenc} 
\usepackage[T1]{fontenc}    
\usepackage{url}            
\usepackage{booktabs}       
\usepackage{amsfonts}       
\usepackage{nicefrac}       
\usepackage{microtype}      
\usepackage{graphicx}
\usepackage{natbib}
\usepackage{doi}

\usepackage{pifont} 
\usepackage{amsmath}
\usepackage{amssymb}
\usepackage{cleveref}       
\usepackage{multirow}
\usepackage{caption}
\usepackage{subcaption}     
\usepackage{makecell}
\usepackage[title]{appendix}
\usepackage[frozencache=true,cachedir=minted-cache]{minted}
\usepackage[disable,bordercolor=white,backgroundcolor=gray!30,linecolor=black,colorinlistoftodos]{todonotes}
\def\codalab{{\em CodaLab Competitions}}
\def\codabench{{\em Codabench}}

\newcommand{\hr}[2]{\href{#1}{\underline{#2}}}
\newcommand{\ra}[1]{\renewcommand{\arraystretch}{#1}}

\newcommand{\quanming}[1]{\todo[inline,color=blue!40]{#1 -- Quanming}}
\title{Codabench: Flexible, Easy-to-Use and Reproducible
Benchmarking Platform}

\date{}

\author{ Zhen Xu \thanks{Zhen Xu is the first author. The other authors are ordered alphabetically.} \\
	4Paradigm, China\\
	\texttt{xuzhen@4paradigm.com} \\
	\And 
	Sergio Escalera \\
	Universitat de Barcelona \\
	Computer Vision Center, Spain \\
	\texttt{sergio@maia.ub.es} \\
	\And 
	Isabelle Guyon \\
	LISN/CNRS/INRIA, Univ. Paris-Saclay, France \\
	ChaLearn, USA \\
	\texttt{guyon@chalearn.org} \\
	\And 
	Adrien Pav\~ao \\
	LISN/CNRS/INRIA, Univ. Paris-Saclay, France\\
	\texttt{adrien.pavao@gmail.com} \\
	\And 
	Magali Richard \\
	Univ. Grenoble Alpes/CNRS, France\\
	\texttt{magali.richard@univ-grenoble-alpes.fr} \\
	\And 
	Wei-Wei Tu \\
	4Paradigm, China \\
	\texttt{tuweiwei@4paradigm.com} \\
	\And
	Quanming Yao \\
	Tsinghua University / 4Paradigm, China \\
	\texttt{qyaoaa@tsinghua.edu.cn} \\
	\And 
	Huan Zhao \\
	4Paradigm, China \\
	\texttt{zhaohuan@4paradigm.com} \\
}

\renewcommand{\headeright}{Submission to Patterns}
\renewcommand{\undertitle}{Submission to Patterns}
\renewcommand{\shorttitle}{Codabench}

\begin{document}
\maketitle

\section*{Highlights}
\begin{itemize}
    \item We developed Codabench, an open-source and community-driven benchmark platform, which facilitates benchmarking and guarantees reproducibility.
    \item  The platform allows organizers to set up benchmarks with custom designs and welcome contributions of users in the form of dataset and/or code submission. 
    \item Organizers may host their own platform instance or use the public instance; in the latter case, they can supply their own compute worker to execute submitted tasks.
    \item Codabench has currently more than 130 users and 2500 submissions. Four use cases in diverse domains are presented in this paper: Graph Machine Learning, Cancer Heterogeneity, Clinical Diagnosis, and Reinforcement Learning
    \quanming{list them, receiving more than 130 users and 2500 submissions} 
    are introduced to demonstrate key features of Codabench.
    
    \item Codabench is amenable, but not limited to, machine learning benchmarking, Artificial Intelligence (AI) for science, and data-centric AI.
\end{itemize}

\section*{Bigger Picture}
In almost all communities working on Data Science,
\quanming{should limit to computer science?} 
researchers face an increasingly severe issue of reproducibility and fair comparison. 
Researchers work on their own version of hardware/software environment, code, data and consequently the published results are hardly comparable. 
We introduce Codabench, which is capable of flexible and easy benchmarking and supports reproducibility. Codabench is an important step towards benchmarking and reproducible research. 
It has been used in various communities including Graph Machine Learning, Cancer Heterogeneity, Clinical Diagnosis and Reinforcement Learning. Codabench is ready to help many trendy researches, 
\quanming{has been used in xxx and is ready to help xxx}
e.g. Artificial Intelligence (AI) for Science and Data-Centric AI.

\section*{In Brief}
Fair and flexible benchmarking is a common issue in data science communities. We develop Codabench platform for flexible, easy and reproducible benchmarking. It is open-source and community-driven. With Codabench, we are able to compare fairly and easily algorithms as well as datasets under diverse protocols. The full reproducibility is guaranteed.

\newpage
\begin{abstract}
Obtaining standardized crowdsourced benchmark of computational methods is a major issue in data science communities. Dedicated frameworks enabling fair benchmarking in a unified environment are yet to be developed. Here we introduce Codabench, an open-source, community-driven platform for benchmarking algorithms or software agents versus datasets or tasks. A public instance of Codabench \url{https://www.codabench.org/} is open to everyone, free of charge, and allows benchmark organizers to compare fairly submissions, under the same setting (software, hardware, data, algorithms), with custom protocols and data formats. Codabench has unique features facilitating the organization of benchmarks flexibly, easily and reproducibly, such as the possibility of re-using templates of benchmarks, and supplying compute resources on-demand. Codabench has been used internally and externally on various applications, receiving more than 130 users and 2500 submissions. As illustrative use cases, we introduce 4 diverse benchmarks covering Graph Machine Learning, Cancer Heterogeneity, Clinical Diagnosis and Reinforcement Learning.
\end{abstract}


\section{Introduction}
\label{sec:introduction}

The methodology of unbiased algorithm evaluation is crucial for machine learning, and has recently received renewed attention in all data science scientific communities. Often, researchers have difficulties understanding which dataset to choose for fair evaluation, with which metrics, under which software/hardware configurations, and on which platform. The concept of benchmark itself is not well standardized and includes many different settings. For instance, the following may be referred to as a benchmark: a set of datasets; a set of artificial tasks; a set of algorithms; one or several dataset(s) coupled with reference baseline algorithms; a package for fast prototyping algorithms for a specific task; a hub for compilation of related algorithm implementations. In addition, many algorithm benchmarks do not offer the easiness to further integrate new methodological developments. A platform for benchmarking tasks in a flexible and reproducible way is thus much needed for everyone to use.

Typical examples of existing frameworks addressing such need are inventoried in Table~\ref{tab:compareplatforms}. Firstly, they include competition platforms, such as Kaggle \citep{kaggle} and Tianchi \citep{tianchi} organizing many data science challenges attracting a large number of participants. They provide elaborate ways of hosting third party competitions and offer services for a fee for commercial competitions. The platform providers retain some control: the organizers do not have full flexibility and control over their competitions. Secondly, data repositories such as UCI repository \citep{Dua:2019} also play an important role for benchmarks and research. But they do not host methods, or results. In contrast, OpenML \citep{OpenML2013} is an example of open-source and free hub of datasets also making available machine learning results. 
However, reproducibility by running code in given containers (or similar ways) is not guaranteed. 
Similarly, PapersWithCode \citep{pwc} collects many tasks and state of the art results from papers. 
But the platform doesn't guarantee the reproducibility of these performances. 
Besides the above mentioned platforms, many domain specific benchmarks exist, 
e.g. DAWNBench \citep{dawnbench}, KITTI Benchmark Suite \citep{KITTI}. 
These benchmarks usually focus on a couple of closely related tasks but are not designed to host general benchmarks. 
In addition, they require repetitive efforts to develop and maintain, which is not always affordable by data science teams. 

Thus,
to facilitate benchmarking, 
we need a platform to allow users to {\bf flexibly and easily} create benchmarks with {\bf custom evaluation protocols and custom data formats}, 
and {\bf execution in a controlled reproducible environment}, 
which is totally {\bf free and open-source}. 
To answer these unmet needs, we developed \codabench. 
A benchmark on \codabench\ consists of one or more flexible tasks (to be explained below) with guaranteed reproducibility.  \codabench\ is the last born of a suite of tools from the open-source ``ChaSuite'' project, receiving over 130 users and 2500 submission on 100 tasks including AutoML, Graph Machine Learning, Reinforcement Learning, detecting cancer heterogeneity and training clinicians. Detailed history and multiple illustrative use cases are introduced in the Appendix. \codabench\ is an important step towards reproducible research and should meet the interest of all areas of data science.


\begin{table}[t]
    \caption{Comparison of various reproducible science platforms.}
    \label{tab:compareplatforms}
    \centering
    \ra{1.3}
    \scalebox{0.75}{
    \begin{tabular}{lp{1cm}<{\centering}p{1.9cm}<{\centering}p{1.2cm}<{\centering}cp{1.4cm}<{\centering}p{1.5cm}<{\centering}p{1.3cm}<{\centering}p{1.35cm}<{\centering}cc}
    \toprule
    \multirow{2}{*}{Platform} & \multicolumn{3}{c}{Flexibility} & \phantom{}& \multicolumn{4}{c}{Easy-to-use} & \phantom{}& \multicolumn{1}{c}{Reproducibility} \\
    \cmidrule[1pt]{2-4} \cmidrule[1pt]{6-9} \cmidrule[1pt]{11-11}
             & Bundle & Result/Code submit & Dataset submit && Easy creation & Open-source/free & API access & Compute queue &&  \\
    \toprule
    Kaggle & \ding{56} & \ding{52} & \ding{56} && \ding{52} & \ding{56} & \ding{52} & \ding{52} && \ding{52}  \\
    Tianchi & \ding{56} & \ding{52} & \ding{56} && \ding{52} & \ding{56} & \ding{56} & \ding{52} && \ding{52}  \\
    UCI & \ding{56} & \ding{56} & \ding{52} && \ding{56} & \ding{52} & \ding{56} & \ding{56} && \ding{52} \\
    OpenML & \ding{56} & \ding{52} & \ding{52} && \ding{52} & \ding{52} & \ding{52} & \ding{56} && \ding{52}  \\
    PapersWithCode & \ding{56} & \ding{52} & \ding{56} && \ding{52} & \ding{52} & \ding{56} & \ding{56} && \ding{52} \\
    DAWNBench & \ding{56} & \ding{52} & \ding{56} && \ding{56} & \ding{52} & \ding{56} & \ding{56} && \ding{52}  \\
    \midrule
    \textbf{Codabench} & \ding{52} & \ding{52} & \ding{52} && \ding{52} & \ding{52} & \ding{52} & \ding{52} && \ding{52} \\ \bottomrule
    \end{tabular}}
\end{table}

\section{Method: Design of Codabench}

\label{sec:design}

\begin{figure}[ht]
        \centering
        \includegraphics[width=0.85\textwidth]{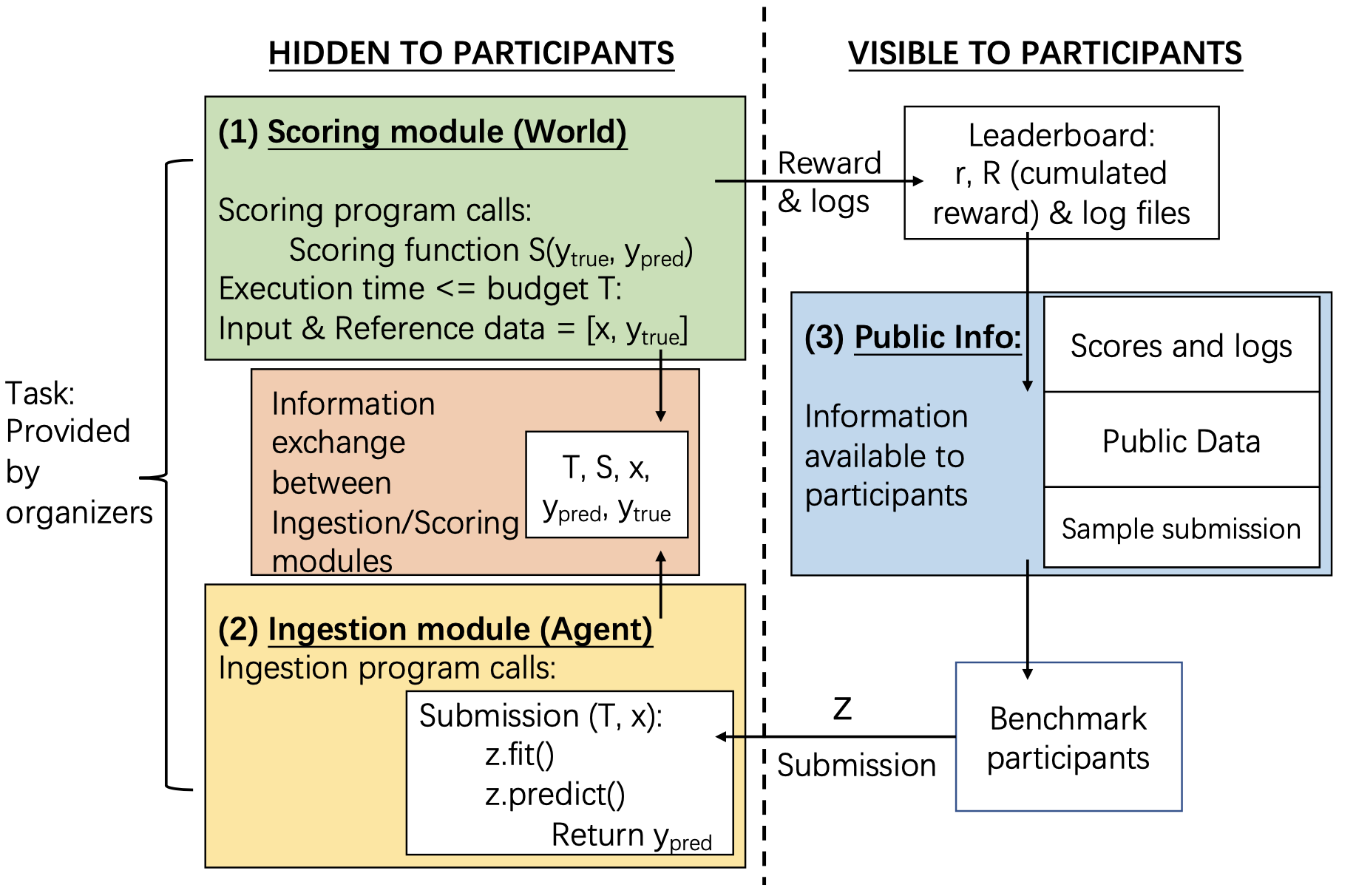}
        \caption[]{{\bf Operational Codabench workflow.} Left side: Task module specified by the organizer, executed on the platform. Right: Web interface with participants permitting to make submissions and retrieve results. Numerated blocks are provided by organizers. They include (1) a scoring module in green; (2)a ingestion module in yellow; (3) and public information in blue. Red block is intermediate information exchange of time budget, scoring, input data, ground truth data and predictions. White bottom right block: participant prepares a submission "z" uploaded to the platform. The submission is then executed by the ingestion program. The role of the scoring program is to produce scores that are then displayed on the leaderboard.}
    \label{fig:internalarchi}
\end{figure}

\codabench\ is a flexible, easy-to-use and reproducible benchmark platform that is open-source and freely provided for everyone. Different from all other platforms, a new concept of task is introduced in \codabench, which is the minimal unit for composing a benchmark. 
A task is composed of an ``ingestion module'' (including ingestion program and input data), 
a ``scoring module'' (including a scoring program and reference data, invisible to the participant's submission), 
a baseline solution with sample data, and meta-data information if needed. Tasks in \codabench\ may be programmed in any programming language in any custom way, which are run in a docker specified by organizers. Figure~\ref{fig:internalarchi} provides a detailed description of \codabench\ internal interaction logistics.



Take supervised learning tasks as an example. A typical usage is that benchmark participants submit a class (e.g. a Python object) ``z'', with 2 methods: \texttt{z.fit} and \texttt{z.predict}, similarly to scikit-learn \citep{scikit-learn} objects.
The ingestion program reads data, calls \texttt{z.fit} with labeled training data and \texttt{z.predict} with unlabeled test data (labeled training data and unlabeled test data being part of the so-called ``input data''), then outputs predictions. The scoring program reads the predictions and evaluates them based on custom scoring metric(s), using the test labels (called ``reference data''). Other application usages are possible, including: transposed benchmarks (datasets are submitted by participants instead of algorithms; the organizers supply a set of algorithms), and reinforcement-learning benchmarks (the ingestion program plays the role of an agent wrapping around the submission of the participant and interacting with a world (scoring program) in a specific way.

The reader is referred to Codabench official repository\footnote{\url{https://github.com/codalab/codabench/}}  where the code and complete documentation are found. In Appendix, we also include instructions and references to get started. To use the public instance of \codabench\, please visit the {\hr{https://www.codabench.org/}{Codabench website}}. To test and install locally, the instructions are given in the readme file of the official repository. The \codabench\ code is released under an {\hr{https://github.com/codalab/codabench/blob/develop/LICENSE}{Apache 2.0 License}}. Under the organization group, there is also \codalab, which is the aforementioned competition platform, and CodaLab Worksheets, which features dynamical workflows, particularly useful for Natural Language Processing. This paper concerns only \codabench.

\section{Results}

\subsection{Key features of Codabench}
\label{sec:features}

\codabench\ is  {\em task-oriented}. Using tasks, 
the organizers have the flexibility of implementing any benchmark protocol, with any dataset format and API,  or even using data generating models,  allowing them to organize reinforcement learning challenges.  In this section, we introduce the key features of \codabench\ contributing to the flexibility, easiness and reproducibility. \codabench\ also supports custom leaderboards and has full {\hr{https://github.com/codalab/codabench/wiki}{documentation}} of usage.

\subsubsection{Flexibility}

Benchmarks are organized by bundles containing all the information of a benchmark. Thanks to the aforementioned task and bundle, \codabench\ supports flexible benchmark types including results submission, code submission and even dataset submission. 


\textbf{Benchmark bundle.}
A benchmark bundle is a zip file containing all necessary constituents of a benchmark: tasks, documentation, and configuration settings (such as leaderboard settings). A \codabench\ bundle may include single or multiple tasks. Classical benchmark is usually single-task while AutoML, Transfer Learning, Meta Learning benchmarks are multi-task.

\textbf{Results or code submission.}
``Classic'' \codabench\  benchmarks are either with result or code submission. On one hand, result submissions are used when organizers wish that participants use they own computational resources. In supervised learning competitions, participants would supply e.g. predictions of output values on some test datasets. Other types of results may be supplied, for instance high resolution images in a hyper-resolution challenge for which inputs are low-resolution images. On the other hand, if the organizers wish to run all algorithms in a uniform manner on the platform, \codabench\ allows the participants to make code submissions. The submitted software is run in a docker supplied by the organizers, either on the default compute worker, or on compute workers supplied by the organizers. This code submission design allows organizers to provide suitable computational resources (e.g. GPUs), and improve reproducibility.

\textbf{Dataset submission.}
To faciliate \textbf{Data-Centric AI}, the role of dataset and algorithm can be transposed with \codabench. In a ``classic'' benchmark, organizers provide dataset(s) and participants submit algorithms. In a transposed benchmark, participants submit datasets and organizers provide reference algorithms. A ``classic'' benchmark will have a leaderboard with datasets in columns, growing by adding more lines are algorithm submissions are made. In a transposed dataset submission benchmark, the leaderboard will have algorithms in columns and lines are added as more datasets are submitted. \codabench\ does not support yet benchmarks in which both dimensions of the leaderboard are grown (i.e. participants can supply either algorithms or datasets).

\subsubsection{Easy-to-use}

A benchmark can be created either with platform editor or by uploading a locally prepared benchmark bundle. Once created, a benchmark can further be modified using the platform editor. An existing benchmark can be saved as another bundle, which facilitates the sharing and portability. Similar benchmark bundles can be easily prepared with shared template bundles. \codabench\ is \textbf{open-source and free} to use.


\textbf{APIs to external clients.} 
We provide {\hr{https://github.com/codalab/codabench/wiki/Robot-submissions}{APIs}} for interacting with the platform, including ``robot'' submissions via command lines, without going through the regular \codabench\ web interface, and likewise a programmatic way of recuperating results directly without going through the leaderboard.

\textbf{Dedicated computing queues.} 
The {\hr{https://codabench.org/}{public instance}} of \codabench\  provides default compute workers. Organizers can also create a dedicated job queue and connect it to their own CPU or GPU compute workers.


\subsubsection{Reproducibility}

\codabench\ makes extensive use of {\hr{https://github.com/codalab/codabench/wiki}{Dockers}}. Benchmark organizers specify the  Docker image by providing its Docker Hub name and tag. It is with Docker that \codabench\ provides full \textbf{reproducibility} to everyone.


\subsubsection{Other features}

\textbf{Custom leaderboard.} To better facilitate benchmarks, the leaderboard is fully customizable and can handle multiple datasets and multiple custom scoring functions. We provide multiple ways to display submissions (best per participant, multiple submissions per participant, etc) and the leaderboard can flexibly ranking submissions by average score, per task, per sub-metric of a certain task, etc.)

\textbf{Documentation.}
The documentation is organized according to stakeholders categories \textit{organizers}, \textit{administrators}, and \textit{contributors} directly on the first page of the documentation \footnote{\url{https://github.com/codalab/codabench/wiki}}. As an \textit{organizer}, you are accompanied with several benchmark templates, from simple to elaborate, to ease the technical aspects of building a benchmark,  and to let you concentrate on scientific aspects of the benchmark.  As an \textit{administrator} of your own instance of \codabench, each piece of the infrastructure is configurable and offered as a docker component. You can deploy your instance in a very flexible way concerning the sizing of your project thanks to deployment guide hints. As an \textit{contributor}, you can discuss with the main developers via the GitHub issues and suggest pull requests to solve some of the issues you have encountered.


\subsection{Use cases of Codabench}
\label{sec:usecases}

\begin{figure}[h]
    \centering
    \begin{subfigure}[b]{0.43\textwidth}
        \centering
        \includegraphics[width=\textwidth]{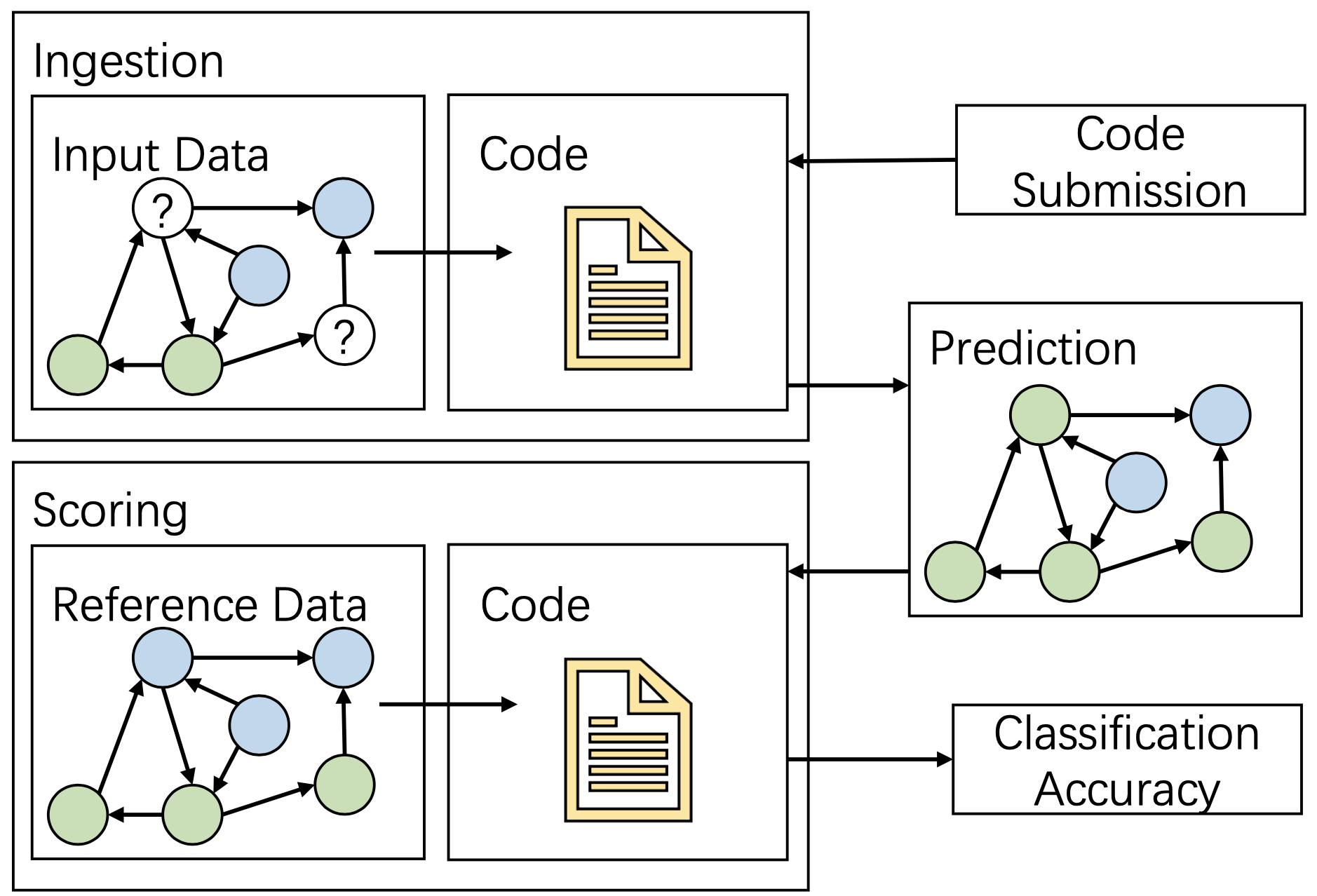}
        \caption{\footnotesize Use case 1: AutoGraph} 
    \end{subfigure}
    \hspace{0.05\textwidth}
    \begin{subfigure}[b]{0.42\textwidth} 
        \centering 
        \includegraphics[width=1\linewidth]{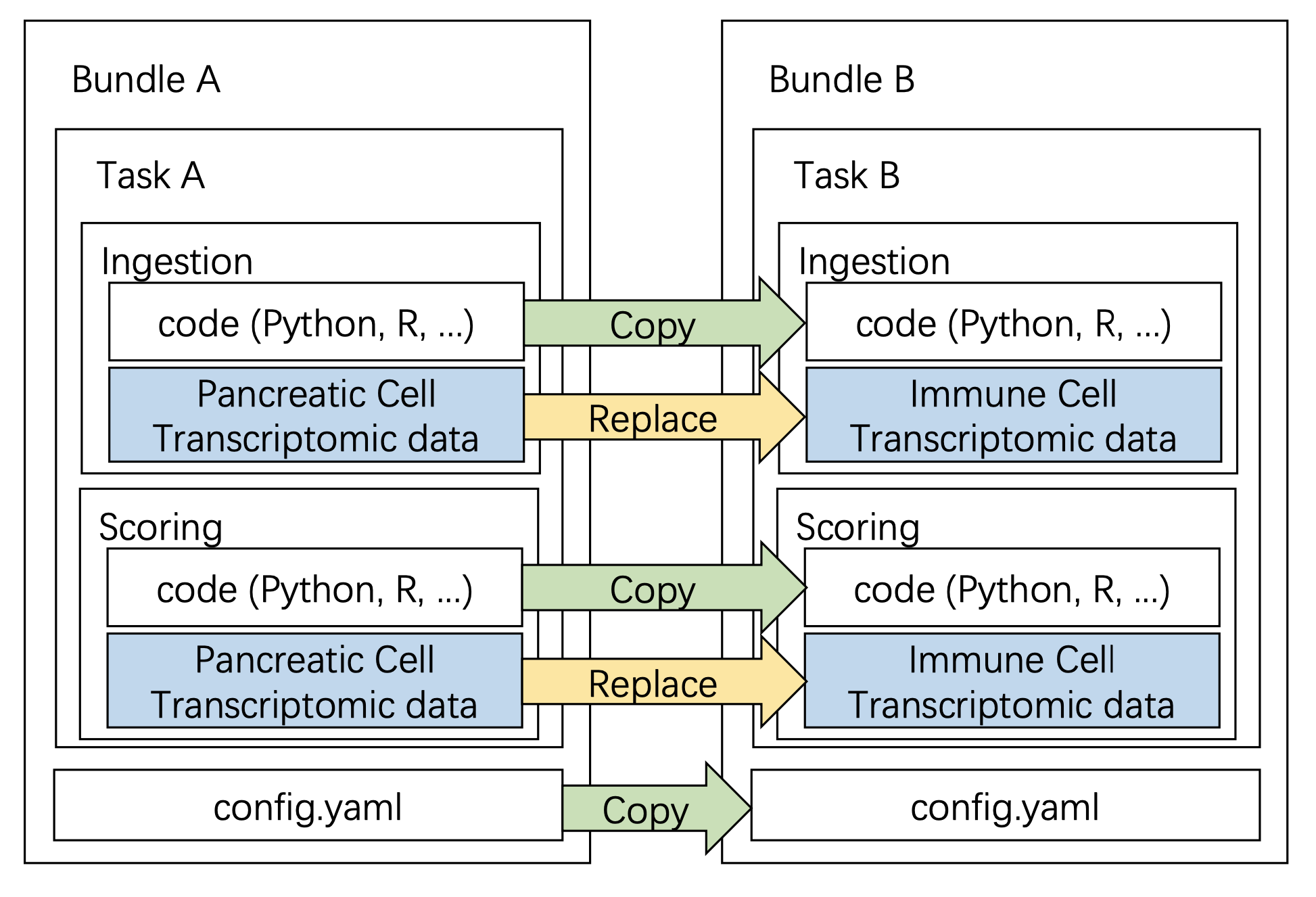}
        \caption{\footnotesize Use case 2: DECONbench}
    \end{subfigure}
    \begin{subfigure}[b]{0.5\textwidth}   
        \centering 
        \includegraphics[width=1\linewidth]{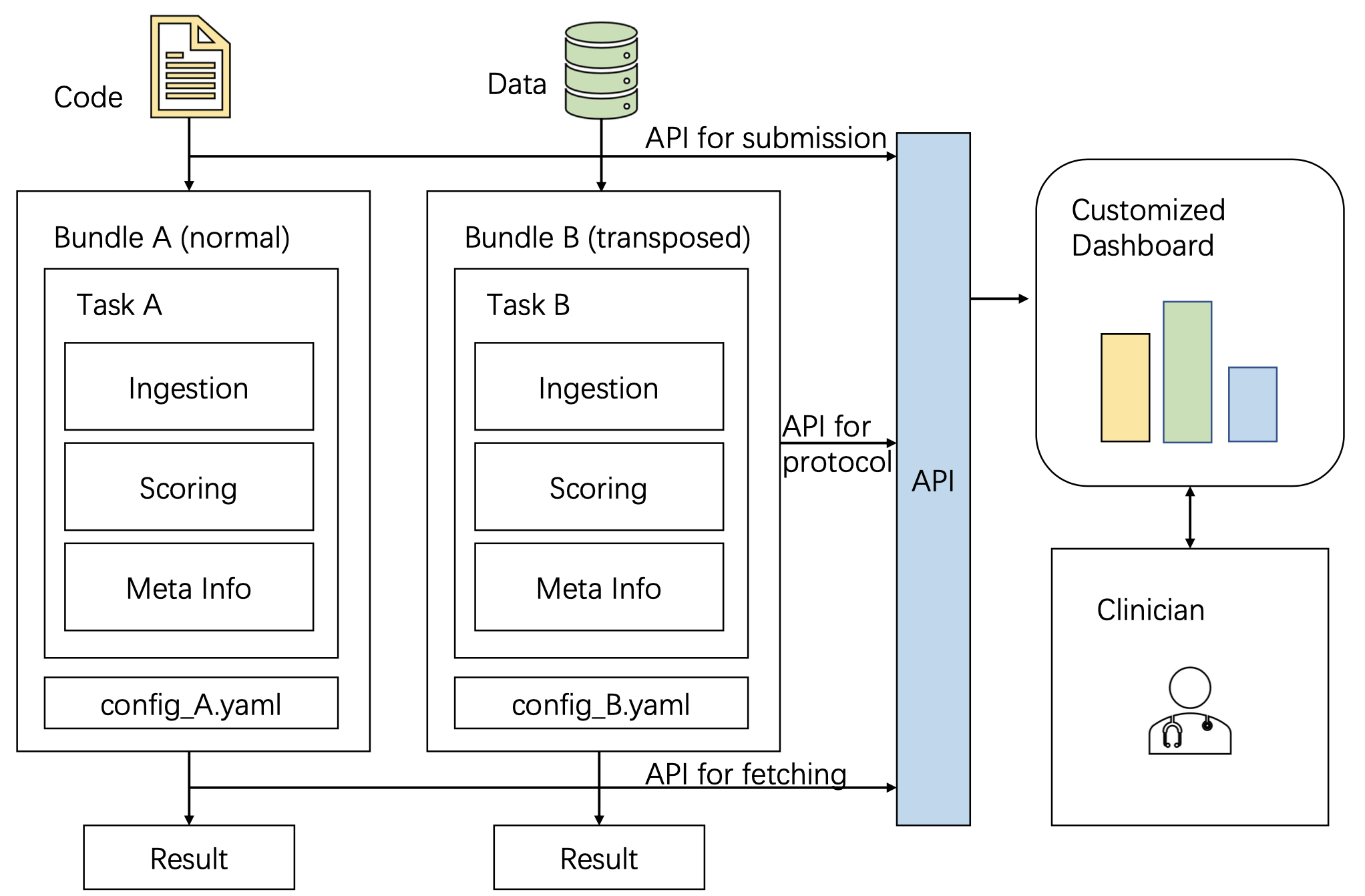}
        \caption{\footnotesize Use case 3: COMETH}
    \end{subfigure}
    \begin{subfigure}[b]{0.45\textwidth} 
        \centering 
        \includegraphics[width=1\linewidth]{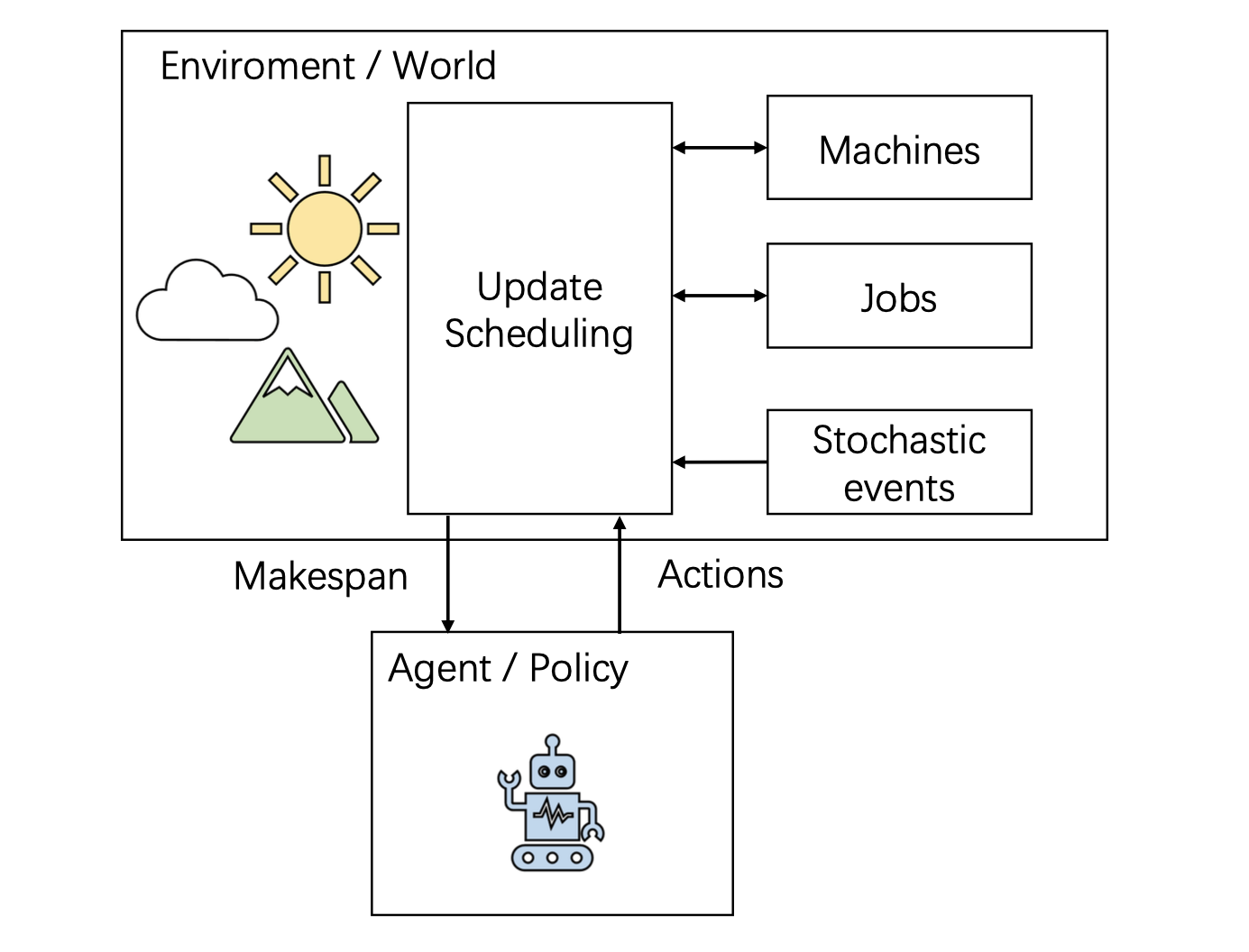}
        \caption{\footnotesize Use case 4: AutoRL}
    \end{subfigure}
    \caption{\bf Use case illustrations.}
    \label{fig:usecases}
\end{figure}

\codabench\ has been used not only internally at 4Paradigm and LISN Lab for tasks of AutoML, Graph Machine Learning, Reinforcement Learning, Speech Recognition and Weakly Supervised Learning, but also externally by University Grenoble Alphes for hosting scientific benchmark in cancer heterogeneity and training clinicians. 
Codabench has received more than 130 users and 2500 submissions distributing on various applications. In this section, we introduce 4 use cases of \codabench, aiming at demonstrating different \codabench\ features and capabilities. A visual illustration is given in Figure \ref{fig:usecases}.

\subsubsection{Use case 1: AutoGraph benchmark}
\label{sec:autograph}

In this section, we introduce Automated Graph Representation Learning (AutoGraph) benchmark, which targets at automated node classification methods on diverse dataset scenarios. With this use case, we show a set of fundamental features of \codabench: \textbf{(1) the code submission mode (2) reproducibility guaranteed by docker (3) flexible benchmark bundle configuration with multiple tasks, and  (4) customizable computational resources.} More technical details can be found in Appendix.

\textbf{Background.} The AutoGraph benchmark inherits from the Automated Graph Representation Learning (AutoGraph) Challenge at  {\hr{https://www.automl.ai/competitions/3\#results}{KDD Cup 2020}}. Graph representation learning has been a very hot topic due to ubiquity of graph-structured data, e.g. social network \citep{surveygnn}, knowledge graph \citep{transe}, etc. The task of our focus here is node classification under the transductive setting.

\textbf{Implementation.} The AutoGraph benchmark is a typical \textbf{code submission} use case. It focuses on AutoML methods which requires more than one dataset to be evaluated together. \codabench\ bundle is by design flexible with \textbf{multiple tasks} each of which contains seperate dataset. We also provide a docker hosted on DockerHub, which will be pulled automatically by \codabench\ platform to run each algorithm submission and could also be used for researchers' local development. Every time a new method is uploaded, a new docker container instance will be called to independently run for each dataset. This way we make sure every algorithm is fairly run under the same setting and the whole process can be \textbf{fully reproduced} on other machines. \codabench\ is designed to adapt to any Docker-enabled computational resource (local machine, cluster server, cloud machines, etc.). We 
currently host the AutoGraph benchmark on \codabench\ with \textbf{free computational resources} thanks to Google's sponsorship, encouraging everyone to contribute\footnote{The public AutoGraph benchmark link will be provided later}. Besides, the datasets are also available to the public for local usage and further benchmarking on \href{https://github.com/NehzUx/AutoGraph-Benchmark}{Github} and \href{https://www.kaggle.com/nehzux/autograph-benchmark}{Kaggle}. To bootstrap the benchmark submissions, we uploaded the solutions of the winners of the challenge. Since the benchmark datasets are released already, users can also run complementary experiments on their local computers and debug mode easily, thus more rapidly making progress. The main incentives to submit to the platform are free hardware and the possibility of showcase results in a common data table.

\subsubsection{Use case 2: DECONbench benchmark}
\label{sec:DECONbench}

In this section, we introduce DECONbench\citep{deconbench} for benchmarking deconvolution methods inferring the tumor micro-environment composition. We show two features of \codabench: \textbf{(1) flexibility of benchmark bundle (in this use case, another task and programming language R supported) (2) reusability and portability of benchmark bundles.} 

\textbf{Background.} Successful treatment of cancer is still a challenge and this is partly due to a wide heterogeneity of cancer composition across patient population. Unfortunately, accounting for such heterogeneity is very difficult and often requires the expertise of anatomical pathologists and radiologists. Therefore, it is pertinent to address this question using computational methods that take advantage of the recent massive generation of high throughput molecular data (called omic data, such as epigenomic or transcriptomic data). DECONbench is a \textbf{series of benchmarks} dedicated to the quantification of intra-tumor heterogeneity on cancer omics data, focusing on estimating cell types and proportion in biological samples using epigenomic and transcriptomic datasets (unimodal and/or multimodal). Participants have to identify an estimation of the cell-types proportion matrix underlying the tumor micro-environment composition. The discriminating metric is mean absolute error (MAE) between prediction and ground truth matrix. Note that DECONbench series is optimized to run methods developed in the statistical \textbf{programming language R}.

\textbf{Implementation.} Using the Codabench platform, the COMETH consortium firstly developed a benchmark for continuous evaluation of computational methods based on epigenomic data\footnote{\url{https://www.codabench.org/competitions/174}}. 
Since we are at the same time interested in other modalities of data under similar task, it would be ideal to reuse previously created bundles instead of going through everything again. Thanks to the portability of \codabench\ bundle design, we only need to replace the data files and adjust slightly the protocol code. All other configuration files can be reused. As a result, this first benchmark was easily cloned and extended to similar benchmarks using other types of data, e.g. all-cell-type transcriptomic data\footnote{\url{https://www.codabench.org/competitions/147}}, immune-cell-types transcriptomic data\footnote{\url{https://www.codabench.org/competitions/148}}, all-cell-types multimodal transcriptomic and epigenomic data\footnote{\url{https://www.codabench.org/competitions/237}}.

\subsubsection{Use case 3: COMETH benchmark}

In this section, we introduce the COMETH benchmark, motivated by real clinical application and it is an exciting step towards Data-Centric AI \footnote{\url{http://datacentricai.org/}}. With this use case, we show that \textbf{(1) \codabench\ supports a transposed benchmark consolidating Data-Centric AI (2) the provided API interaction opens a window for other customization scenarios.} 

\textbf{Background.} When it comes to clinical application, it is often necessary for health data scientists and clinicians to identify the most suitable existing method to be applied on a given dataset. In this case, we focus more on the data used for training and inference instead of algorithmic development, which aligns with Data-Centric AI. 

\textbf{Implementation.} To solve this question, the COMETH consortium developed the COMETH benchmark
\footnote{\url{https://www.codabench.org/competitions/218}},
a transposed challenge in which datasets should be submitted to be evaluated against existing different reference deconvolution methods (ie “tasks” in the Codabench design) and people can retrieve the corresponding outputs, in a fully reproducible environment. To facilitate the use of this functionality by clinicians who are less familar with data science programming details, COMETH benchmark has been connected to an external client displaying a user-friendly web dashboard. This external client is able to send requests to users directly on the COMETH benchmark using APIs provided by \codabench\ and return the generated results from all reference algorithms. This feature strongly contributes to a direct transfer of knowledge between data scientists and healthcare professionals. This design was used at a winter school for training clinicians and data scientists \footnote{\url{https://cancer-heterogeneity.github.io/cometh.html}}.

\subsubsection{Use case 4: AutoRL benchmark}

We lastly introduce another use case: AutoRL benchmark focusing on reinforcement learning and operational research. With this use case, we show that \codabench\ is \textbf{RL-friendly} with the help of flexible design of benchmark bundles.

\textbf{Background.} We consider the problem of Dynamic Job-Shop Scheduling. The task is to allocate a set of jobs to a set of machines with stochastic events. Each job has a pre-determined operation sequence to be executed on certain machines. To mimic real life scenarios, we add aleatoric machine down events to the problem. We thus expect an agent policy making decisions on how to schedule better the jobs in minimal time. The reward depends on the makespan.

\textbf{Implementation.} This task can be well formulated into a reinforcement learning framework. As explained in Sec 2, our design of bundle and ingestion/scoring program makes it very natural and flexible for RL problems. We could either follow Figure \ref{fig:internalarchi} and use scoring as environment and ingestion as agent, or it is also possible to wrap everything into the ingestion module. 

\section{Discussion and Conclusion}
\label{sec:conclusion}

\codabench\ is a new open-source platform for data science benchmarks. \codabench\ is compatible with diverse tasks (including supervised learning and reinforcement learning) and and supports result, code, and dataset submission. It is easy to use \codabench\ and reproducibility is guaranteed by Dockers. \codabench\ has a public instance free for use, deployed at Université Paris-Saclay, but can also be deployed locally, with the technology stack provided in documentation. Hosting, maintaining, and further developing the platform is funded by grants and donations. As real scenarios, we introduce 4 benchmark use cases illustrating the flexibility, easiness in use, reproducibility and other features of \codabench. We note also that tremendous other tasks could be integrated into \codabench\ as well including EEG classification, drug discovery and property prediction, dynamic simulation for weather, traffic, fluid, etc., which are important tasks towards AI for Science.

The current limitations of \codabench\ are mainly as follows. First, since it is relatively new, we do not have yet an active community of organizers and benchmark participants. Second, although supported by design, we have not had yet a distributed computation scenario, where complex multi-node compute workers are used. This could enrich our benchmark template library with benchmarks for algorithm parallelization. Thirdly, although \codabench\ supports both code submission and dataset submission, we do not currently allow users to extend the leaderboard in both directions simultaneously, i.e. submit either code or datasets. This feature could largely increase the user experience of the platform. Lastly, \codabench\ doesn't support yet hardware related benchmarks or human-in-the-loop benchmarks which could be interesting to consider in the future.

Potentially harmful uses of \codabench\ could result from poor benchmark designs (e.g. no scientific question is asked by hosting a benchmark), or bad data collections (e.g. data license, data quality), as in any machine learning project. We are working on an open-access book (to appear in 2022) on best practices for designing challenges and benchmarks including data preparation, task evaluation,  benchmark analyse paper, etc.

Further work includes providing more comprehensive usage templates illustrating features such as: (1) splitting an algorithm workflow into sub-modules and scoring the effectiveness of the modules individually (e.g., with ablation or sensitivity analysis); (2) providing templates of fact sheets to extract information about algorithms (similar to datasheets for datasets, but for algorithms); and (3) providing guidelines to benchmark participants to produce enriched detailed results, amenable to meta-analyses.

\section*{Acknowledgments}
\label{sec:acknowledgement}
The Codabench\ project shares the same  \hr{https://github.com/codalab/codalab-competitions/blob/master/docs/Community-Governance.md}{community governance} as \codalab. 
The openness of \codabench\ is total: the \hr{https://github.com/codalab/codabench/blob/develop/LICENSE}{Apache 2.0 licence} is used, the \hr{https://github.com/codalab/codabench}{source code is on GitHub}; the development framework and all the used components are open-source. 
 \codabench\  has received important contributions from many people who did not co-author this paper, and we would like to thank their efforts in making \codabench\  what it is today, including early \codalab\ developers and contributors (alphabetically):
Pujun Bhatnagar, 
Justin Carden,  
Richard Caruana, 
Francis Cleary, 
Xiawei Guo, 
Ivan Judson, 
Lori Ada Kilty, 
Shaunak Kishore, 
Stephen Koo,
Percy Liang, 
Zhengying Liu,
Pragnya Maduskar, 
Simon Mercer, 
Arthur Pesah, 
Christophe Poulain, 
Lukasz Romaszko, 
Laurent Senta, 
Lisheng Sun,
Sebastien Treguer
Cedric Vachaudez, 
Evelyne Viegas, 
Paul Viola, 
Erick Watson, 
Tony Yang, 
Flavio Zhingri, 
Michael Zyskowski. 
We would like thank particularly the people who contributed to the design, development, and testing of \codabench\ including (alphabetically):
Alexis Arnaud,
Xavier Bar\'o,
Feng Bin,
Yuna Blum,
Eric Carmichael,
Laurent Darré.
Hugo Jair Escalante,
Sergio Escalera,
Eric Frichot,
Yuxuan He,
James Keith,
Anne-Catherine Letournel,
Shouxiang Liu,
Zhenwu Liu,
Adrien Pavao,
Magali Richard,
Tyler Thomas,
Nic Threfts, 
Bailey Trefts,
Catherine Wallez,
Lanning Wei.
Université Paris-Saclay is hosting the main instance of \codabench. Funding and support have been received by several research grants, including Big Data Chair of Excellence FDS Paris-Saclay, Paris Région Ile-de-France, EU EIT projects HADACA and COMETH, United Health Foundation INCITE project, and ANR Chair of Artificial Intelligence HUMANIA ANR-19-CHIA-0022, 4Paradigm, ChaLearn, Microsoft, Google. We also appreciate the following people and institutes for open sourcing datasets which are used in our use cases: Andrew McCallum, C. Lee Giles, Ken Lang, Tom Mitchell, William L. Hamilton, Maximilian Mumme, Oleksandr Shchur,  David D. Lewis, William Hersh, Just Research and Carnegie Mellon University, NEC Research Institute, Carnegie Mellon University, Stanford University, Technical University of Munich, AT\&T Labs, Oregon Health Sciences University. This work has been partially supported by the Spanish project PID2019-105093GB-I00 and by ICREA under the ICREA Academia programme. We are also very grateful to Joaquin Vanschoren for fruitful discussions.

\bibliographystyle{unsrtnat}
\bibliography{ref}  

\newpage
\begin{appendices}

\section{ChaSuite family and comparison between competition/benchmark}

\begin{figure}
        \centering
        \includegraphics[width=0.9\textwidth]{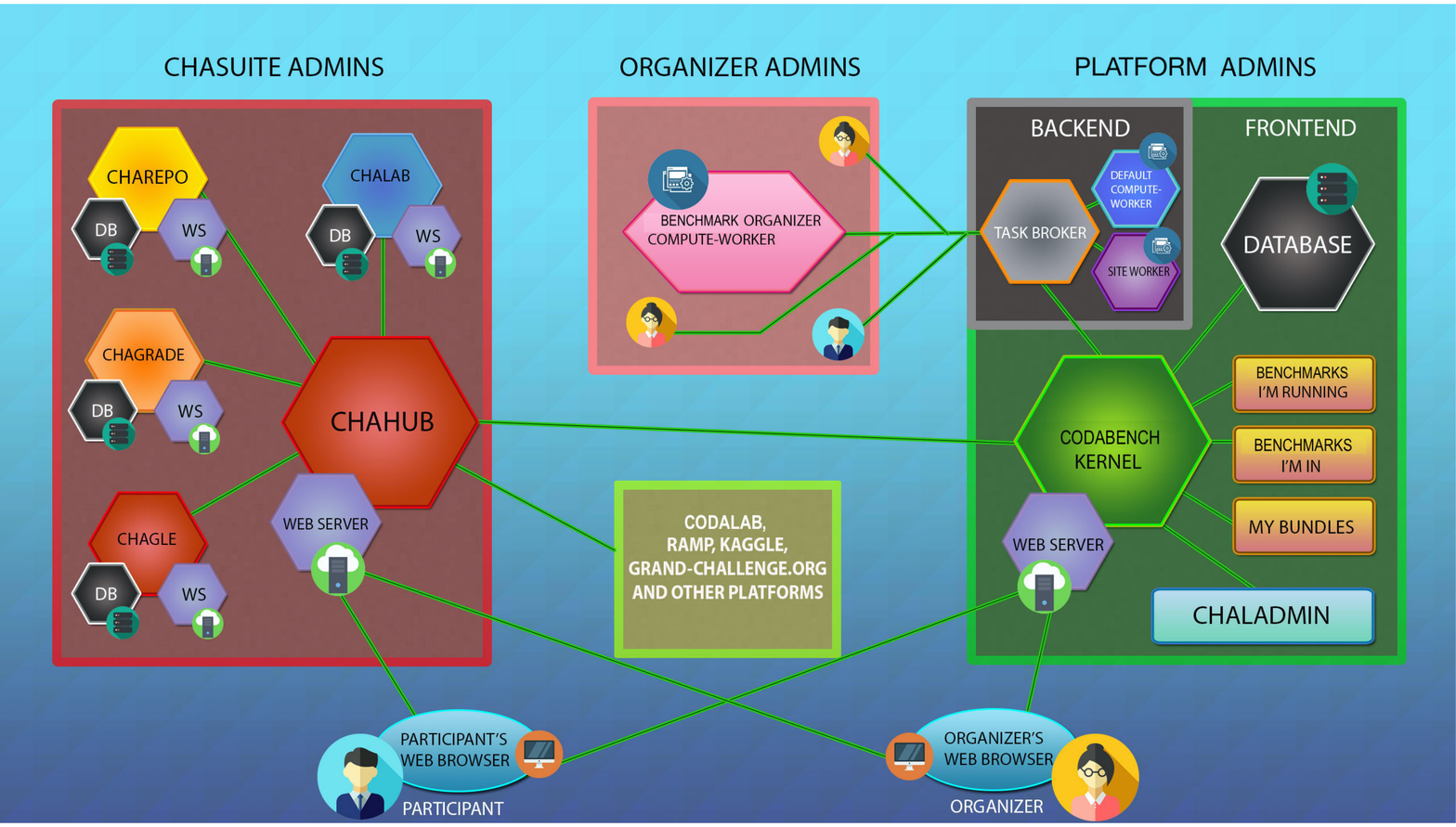}

        \caption[]{{\bf ChaSuite architecture.} \codabench\ is part of the \codalab\ project, including a suite of tools to organize challenges and benchmarks called the``ChaSuite". Right: The kernel of Codabench is interfaced with a web browsers, a database, and a backend dispatching jobs to compute-workers, configured and administered by organizers. Left: The ChaSuite includes an index of competitions and benchmarks (ChaHub) with a search engine (Chagle), a wizard to design challenges (ChaLab), a data repository (ChaRepo), a tool to administer classes (ChaGrade).}
    \label{fig:generalarchi}
\end{figure}

\codabench\ is the last born of a suite of tools from the open-source ``ChaSuite'' project (Figure~\ref{fig:generalarchi}), which all have public instances available for free use of charge. ``ChaSuite'' provides a comprehensive suite of tools for competition and benchmark organizers. \codabench\ is inspired by \codalab, an open-source platform for running data science competitions, which has been used in hundreds of challenges associated to physics, machine learning, computer vision, natural language processing, health and life sciences, among many other fields. Data science competitions have played an important role for solving machine learning problems both in theory and application (e.g. ImageNet challenge \citep{ILSVRC15}, the Netflix Prize \citep{netflix}, etc). Benchmarks can be viewed as a never-ending competition enabling continuous evaluation of methods under the same settings (see Table~\ref{tab:diffcompbench} in the Appendix for a comparison between benchmark and competitions).

Compared with \codalab, \codabench\ has made significant improvements to better address the organization of benchmarks. The full code has been completely rewritten and the code base is much cleaner and maintainable. We introduce a new ``task'' concept (as mentioned in Sec 2 and Sec 3) for flexibility and portability purposes. We now support data submission in addition to results and code submission, which makes \codabench\ an important platform for Data Centric AI, which is a new trending paradigm focusing more on the underlying data used to train and evaluate models. We also provide low level APIs to facilitate third party's customization. A new fact sheet system has been added to allow submit more information in an integrated way and the leaderboard now supports multiple modes of display and advanced ranking. 

\textbf{Backward compatibility with Codalab}. While \codabench's novelty is the possibility of creating benchmarks, it is fully compatible with \codalab. Competition bundles in the old format e.g. dumped from \hr{http://competitions.codalab.org}{the Codalab public instance} can be re-uploaded to \codabench. Competition features such as having multiple-phases (not usually relevance for benchmarks) are supported for compatibility reasons in \codabench. Multi-phase challenges help organizers keep participants engaged over long periods of time.

\begin{table}[ht]
    \caption{Comparisons of competition and benchmark.}
    \label{tab:diffcompbench}
    \centering
    \scalebox{0.9}{
    \begin{tabular}{p{2.5cm}|p{5.5cm}|p{6cm}}
    \toprule
         & Competition & Benchmark  \\
    \midrule
        Purpose & Crowdsourcing problems in a short time and harvesting solutions & Continuous fair evaluation, over a long time period, in a unified framework  \\
    \midrule
        Phases & Multiple phases & Single phase  \\
    \midrule
        Time period & Usually limited & Often never ending  \\
  \midrule
          Cooperation \& information sharing & Limited due to the competitive nature &  As extensive as possible \\
    \midrule
        Submissions & Usually algorithm predictions or algorithm code &  Algorithm code or datasets; code or dataset name, description, documentation meta-data and/or fact-sheets; scoring programs for custom analyses \\
    \midrule
        Outcome & Leaderboard with usually a single global ranking based on one score from each team (last or best) & Table with all the submissions made; sorting with multiple scores possible; multiple analyses, graphs, figures, code sharing \\
    \bottomrule
    \end{tabular}
    }
\end{table}

\section{Codabench usage: getting started}
\label{sec:gettingstarted}

Using \codabench\ as a participant is straightforward. First, create an account and login on {\hr{https://www.codabench.org/}{\codabench}}. Then choose an existing benchmark to join following the instructions provided by the organizers. To organize a benchmark, a user can either use the \codabench\ editor or upload a benchmark bundle which is a zip file containing code, dataset, and configuration file. Detailed instructions are found on {\hr{https://github.com/codalab/codabench/wiki}{\codabench\ Documentation}}. For advanced users who wish to deploy a private instance of \codabench\, please refer to \codabench\ deployment instructions in the same wiki. To illustrate better the benchmark bundle, we provide a simplified bundle example in 
the next section, which contains ingestion program, scoring program, data, text descriptions and a configuration YAML file.

\section{Sample bundle file for Codabench}
\begin{figure}[H]
        \centering
        \includegraphics[width=0.6\textwidth]{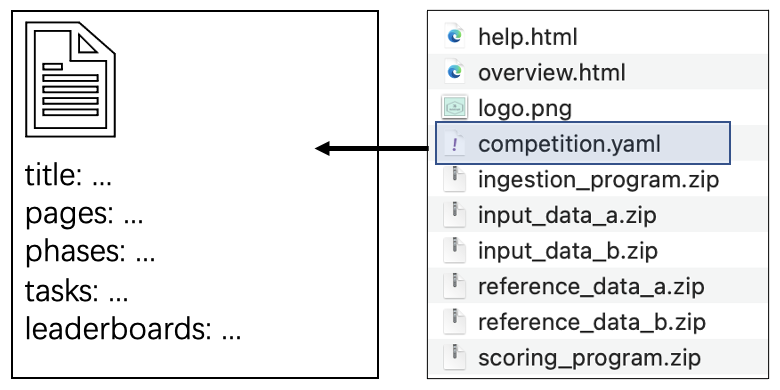}
        \caption[]{{\bf Bundle structure.} The details of \texttt{competitions.yaml} is given below.}
    \label{fig:bundle}
\end{figure}

In this section, we provide a concrete bundle example to show how simple it is to organize benchmarks on \codabench. A bundle consists of five parts as in Figure \ref{fig:bundle}: (1) a YAML configuration file (2) ingestion program (3) scoring program (4) data (5) text files for additional description.

The \textbf{ingestion program} usually reads data and participant's submission. It calls participant's method on the dataset and produces predictions to a shared space. The \textbf{scoring program} usually reads ingestion program's output and evaluate w.r.t ground truth according to organizer customized metric. It finally writes scores to a text file which will be read by platform and be displayed on leaderboard. The \textbf{data} contain input data (in supervised learning,  they are usually X\_train, y\_train, and X\_test) and reference data (in supervised learning,  it is usually y\_test). Both are zipped into separate files. The \textbf{text files} are just html or markdown files for organizers to provide other information e.g. instructions, references, etc. A final \textbf{YAML file} connects all previous parts and provides more configurations for the benchmark. A simplified YAML file is as follows. It contains general configurations like title, logo image, docker image, and  which htmls to be displayed, leaderboard configuration (e.g. which metrics will be used in the leaderboard) and tasks. Each task is by itself a complete unit for running. It contains name, id, ingestion program, scoring program, input data, reference data.

\newpage
\begin{minted}[
    gobble=4,
    frame=single,
    linenos
  ]{YAML}
    # Sample YAML file based on AutoGraph benchmark
    title: 'AutoGraph Benchmark'
    description: 'Automated Graph Representation Learning Challenge'
    docker_image: nehzux/kddcup2020:v2 # Docker Hub ID
    pages:                    # These are "free style" documentation pages
        - title: help         # You can have any title and file name
            file: 'help.html' # You may use HTML or Markdown (.md files)
        - title: overview     # These pages will show up in the benchmark site
            file: 'overview.html'
    phases:                   # Benchmarks usually have s single phase 
                              # (competitions may have several) 
        - index: 0            # Phase order number   
            name: 'AutoGraph'
            start: 2021-01-01
            end:   2022-12-31
            tasks:            # Tasks included in this phase
                - 0           # Reference number in task list below,
                - 1           # or absolute reference in Codabench database
            max_submissions: 1000
            max_submissions_per_day: 100
            execution_time_limit_ms: 60000
    tasks:                    # Tasks for the above defined phase
        - index: 0
            name: 'Task a'    # For public display on leaderboad
            description: 'Dataset a'   # Private comments
            # Ingestion module:
            ingestion_program: ingestion_program.zip
            input_data: input_data_a.zip
            # Scoring module
            scoring_program: scoring_program.zip
            reference_data: reference_data_a.zip
            # whether the ingestion program is run first, then the 
            # scoring program, or the are run in parallel
            ingestion_only_during_scoring: True
        - index: 1
            name: 'Task b'
            description: 'Dataset b'
            # Ingestion module:
            ingestion_program: ingestion_program.zip
            input_data: input_data_b.zip
            # Scoring module            
            reference_data: reference_data_b.zip
            scoring_program: scoring_program.zip
            ingestion_only_during_scoring: True
    leaderboards:               # Leader board form 
        - title: Results        # single leaderboard supported in this version
            key: main           # main key, leave untouched
            columns:
            - title: 'Acc'      # Name of the column displayed
                key: acc        # Data key name used by scoring program
                index: 0        # Order of columns
                sorting: desc   # Sort in descending order
            - title: 'BalAcc'
                key: bacc
                index: 1
                sorting: desc
\end{minted}

\end{appendices}
\end{document}